\pdfoutput=1

\documentclass[11pt]{article}

\usepackage[final]{acl}

\usepackage{times}
\usepackage{latexsym}

\usepackage[LGR,T1]{fontenc}

\usepackage[utf8]{inputenc}

\usepackage{microtype}

\usepackage[greek.polutoniko, english]{babel}

\usepackage{substitutefont}
\substitutefont{LGR}{\rmdefault}{cmr}

\usepackage{graphicx}
\usepackage{booktabs}
\usepackage{threeparttable}

\title{Sentence Embedding Models for Ancient Greek Using Multilingual Knowledge Distillation}

\author{Kevin Krahn \and  Derrick Tate \and Andrew C. Lamicela \\
        Sattler College, Boston, MA \\ \texttt{\{kevin.krahn24, dtate, alamicela\}@sattler.edu}}

\begin{document}
\maketitle
\begin{abstract}
Contextual language models have been trained on Classical languages, including Ancient Greek and Latin,
for tasks such as lemmatization, morphological tagging, part of speech tagging, authorship attribution,
and detection of scribal errors.
However, high-quality sentence embedding models for these historical languages are significantly more difficult
to achieve due to the lack of training data. In this work, we use a multilingual knowledge distillation approach to train
BERT models to produce sentence embeddings for Ancient Greek text. The state-of-the-art sentence
embedding approaches for high-resource languages use massive datasets, but our distillation approach allows
our Ancient Greek models to inherit the properties of these models while using a relatively small amount of
translated sentence data. We build a parallel sentence dataset using a sentence-embedding alignment
method to align Ancient Greek documents with English translations, and use this
dataset to train our models.
We evaluate our models on translation search, semantic similarity, and semantic retrieval tasks and investigate
translation bias.
We make our training and evaluation datasets freely available at \href{https://github.com/TickleForce/ancient-greek-datasets}{this url}.
\end{abstract}

\section{Introduction}

Sentence embedding models, which map sentences or other sequences of text to a dense vector space, such that
semantically similar sentences are close together in the vector space, have many applications
in NLP. Current state-of-the-art sentence embedding models, however, are trained on modern, high-resource languages
such as English and use massive datasets consisting of billions of sentence pairs \cite{ni-etal-2022-sentence}. A different
approach is needed for historical languages, which have much less data available.

In this work, we train several sentence embedding models for Ancient Greek.
Many more Ancient Greek texts have survived compared to texts from most other
ancient languages, which makes sentence embedding models both more feasible and useful.

Several previous works have trained language models for Ancient Greek. \citet{johnson-etal-2021-classical}
introduced the Classical Language Toolkit (CLTK) which includes several tools for Ancient Greek processing,
including static word embeddings. \citet{singh-etal-2021-pilot} fine-tuned a Modern Greek BERT model
\citep{koutsikakisGREEKBERTGreeksVisiting2020} on Ancient Greek text for PoS tagging, morphological tagging, and lemmatization tasks. \citet{yamshchikov-etal-2022-bert}
trained a BERT model for authorship classification of Pseudo-Plutarch texts. \citet{cowen-breenLogionMachineLearning2023}
trained another BERT model for the purpose of identifying errors in scribal transmission.
\citet{riemenschneiderExploringLargeLanguage2023} produced the most comprehensive work on
Classical language models to date, training multiple models on a large multilingual corpus of
Ancient Greek, Latin, and English texts and comprehensively evaluating and comparing
their new models to previous models on a variety of tasks. None of these works, however,
produce sentence embedding models for Ancient Greek.

Although there are many digitized Ancient Greek texts available, there is a lack of suitable training
data for training sentence embedding models from scratch. The best approaches for high-resource languages
involve large human-annotated datasets, such as the natural language inference (NLI) datasets used
by Sentence-BERT \citep{reimers-gurevych-2019-sentence}.
Needless to say, such datasets are not available for Ancient Greek.

\begin{figure}[ht]
	\centering
	\includegraphics[width=7cm]{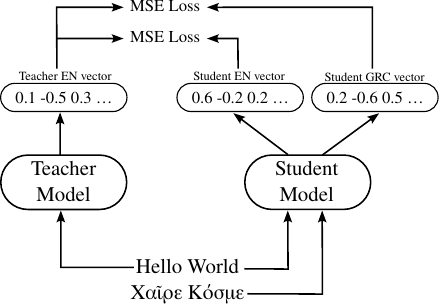}
	\caption{Multilingual knowledge distillation for English to Ancient Greek sentence pairs.}
	\label{fig:fig1}
\end{figure}

Following \citet{reimers-gurevych-2020-making}, we use \textit{multilingual knowledge distillation}
to train sentence embedding models with an aligned vector space for Ancient Greek and English.
Given a teacher model \(M\) for a language \(s\), and a dataset
of translated sentences \(((s_{1},t_{1})..(s_{n},t_{n}))\) where \(s_{i}\) and \(t_{i}\) are parallel sentences,
we train a new student model \(\hat{M}\) to mimic the sentence embeddings of the teacher \(M\) using mean squared loss,
such that \(\hat{M}(s_i) \approx M(t_i)\) and \(\hat{M}(t_i) \approx M(s_i)\). In our case,
the teacher model is English and the student model learns both
Greek\footnote{When we refer to ``Greek'' in an unqualified way in this paper we are referring to Ancient Greek.}
and English embeddings.

This approach has numerous advantages:
1) it requires a relatively small amount of training data,
2) the student model inherits the vector space properties of a state-of-the-art English sentence embedding model,
3) the student model is multilingual, and
4) the vector spaces are aligned across languages.

The cross-lingual nature of this approach is especially useful for Ancient Greek semantic retrieval,
since it is much easier to formulate search queries in English than in Ancient Greek.
Although it is possible to operate on the English translations of Greek texts,
translations are not readily available for all Greek texts, and the available translations are
usually not aligned at the sentence level, making it difficult to quickly
find the corresponding Greek text.
Furthermore, English translations can suffer from various kinds of translator bias,
whereas a language model that operates directly on the Greek text can offer an ``average'' of multiple translators'
interpretations of the text (See Section \ref{sec:bias}).

We produce a training dataset of parallel sentences using a two-step translation alignment process:
an initial, smaller dataset was produced using a
sentence-length heuristic and dictionary-based alignment technique \citep{halacsyParallelCorporaMedium2007},
and this initial dataset was used to train an intermediate multilingual sentence embedding model, which was used to align a larger dataset
using the approach introduced by \citet{liuBertalignImprovedWord2023}, which uses
sentence embeddings for state-of-the-art alignment quality.

We create new evaluation datasets for Ancient Greek translation search,
semantic textual similarity (STS), and semantic retrieval (SR)
and we evaluate our models on these datasets.

In summary, our contributions are as follows:
\begin{enumerate}
  \item We use a multilingual knowledge distillation approach to train several Ancient Greek sentence embedding models.
  \item We use translation alignment to produce a dataset of Ancient Greek sentences and their English translations.
  \item We develop evaluation datasets for translation search, semantic retrieval, and semantic textual similarity, and we evaluate
  our sentence embedding models on these tasks.
\end{enumerate}

\section{Training}

\subsection{Base Models}

To train a sentence embedding model, we first need a base language model trained on Ancient Greek text.
The existing Ancient Greek language models were unsuitable for our purposes;
most of them are monolingual, but we are training a multilingual model.
The models trained by \citet{riemenschneiderExploringLargeLanguage2023} would be
the best candidates because they include English, but one of their goals was to avoid contamination from modern
languages, such as modern concepts and technology like cellphones which were unknown in antiquity.
However, for us this is not a concern, since one of our goals is to train a model to facilitate
semantic search with modern language and terminology.

Instead, we fine-tune multilingual BERT-base (mBERT) \citep{devlin-etal-2019-bert}
and XLM-RoBERTa-base (XLM-R) \citep{conneau-etal-2020-unsupervised} for our base models.
\citet{pires-etal-2019-multilingual} shows that low-resource languages can benefit
from multilingual pre-training. We use masked language modeling (MLM) to fine-tune mBERT,
(denoted as \textsubscript{GRC}mBERT) and XLM-R (denoted as \textsubscript{GRC}XLM-R)
with Ancient Greek text, and we use these as base models. See Appendix \ref{sec:appendix_a} for training details.

Both mBERT and XLM-R were trained on Modern Greek,
among many other languages, but not on Ancient Greek, and hence one disadvantage of these models
is that their tokenizers are not optimized for Ancient Greek morphology, which could
negatively impact performance \citep{park-etal-2021-morphology,hofmann-etal-2021-superbizarre}.

We use a similar approach to \citet{yamshchikov-etal-2022-bert} to compare the
mBERT and XLM-R tokenizers.
We take a random sample of 20k Ancient Greek sentences from the pre-training
corpus and compute the average token length and average words
per token for a rough estimation of tokenization quality (See Table \ref{table:tokenizers}).
The XLM-R tokenizer scores higher on both metrics compared to the mBERT tokenizer.
However, a higher score for either metric does not guarantee superior performance
in downstream tasks, since it does not measure how well the sub-word tokens
capture Ancient Greek morphology.

\begin{table}[t]
	\centering
	\begin{tabular}{lrr}
		\textbf{Model} & \textbf{Symbols/token} & \textbf{Words/token} \\
		\midrule
		mBERT & 2.29 & 0.37 \\
		XLM-R & 2.66 & 0.43 \\
		\bottomrule
	\end{tabular}
	\caption{The XLM-R tokenizer produces longer tokens and a higher number of words per
	token on Ancient Greek text compared to the mBERT tokenizer.}
	\label{table:tokenizers}
\end{table}

\subsection{Knowledge Distillation}

To train multilingual sentence embedding models on English and Ancient Greek with an aligned vector space
we use \textit{multilingual knowledge distillation} \citep{reimers-gurevych-2020-making}.
This process requires a teacher model \(M\) for a source language \(s\), and a dataset
of translated sentences \(((s_{1},t_{1})..(s_{n},t_{n}))\) where \(s_{i}\) and \(t_{i}\) are parallel sentences.
We train a student model \(\hat{M}\) to mimic the sentence embeddings of the teacher \(M\)
such that \(\hat{M}(s_i) \approx M(t_i)\) and \(\hat{M}(t_i) \approx M(s_i)\).
The following mean squared loss function is minimized for each mini-batch $\beta$:
\[
\resizebox{\columnwidth}{!}{
	$
	\displaystyle
	\frac{1}{|\beta|}\sum_{j\in\beta}\left[((M(s_{j})-\hat{M}(s_j))^2 + (M(s_{j})-\hat{M}(t_j))^2\right]
	$}
\]
Thus, the student $\hat{M}$ learns to map each target and source sentence to the same location
in vector space.

For the teacher $M$ we compare two models:
\begin{enumerate}
  \item \texttt{all-mpnet-base-v2},\footnote{\url{https://huggingface.co/sentence-transformers/all-mpnet-base-v2}}
  a model tuned for semantic search, trained on a large and diverse training set of 1B+ pairs
  (Denoted as \texttt{mpnet}).
  \item \texttt{sentence-t5-large},\footnote{\url{https://huggingface.co/sentence-transformers/sentence-t5-large}}
  a T5 model tuned for sentence similarity tasks, trained on 2B pairs \citep{ni-etal-2022-sentence}
  (Denoted as \texttt{st5}).
\end{enumerate}

Both above models have a final normalization layer which we remove prior to training
to allow student model to learn the original vector space properties of the
teacher model.

\begin{figure}[ht]
	\centering
	\includegraphics[width=7.7cm]{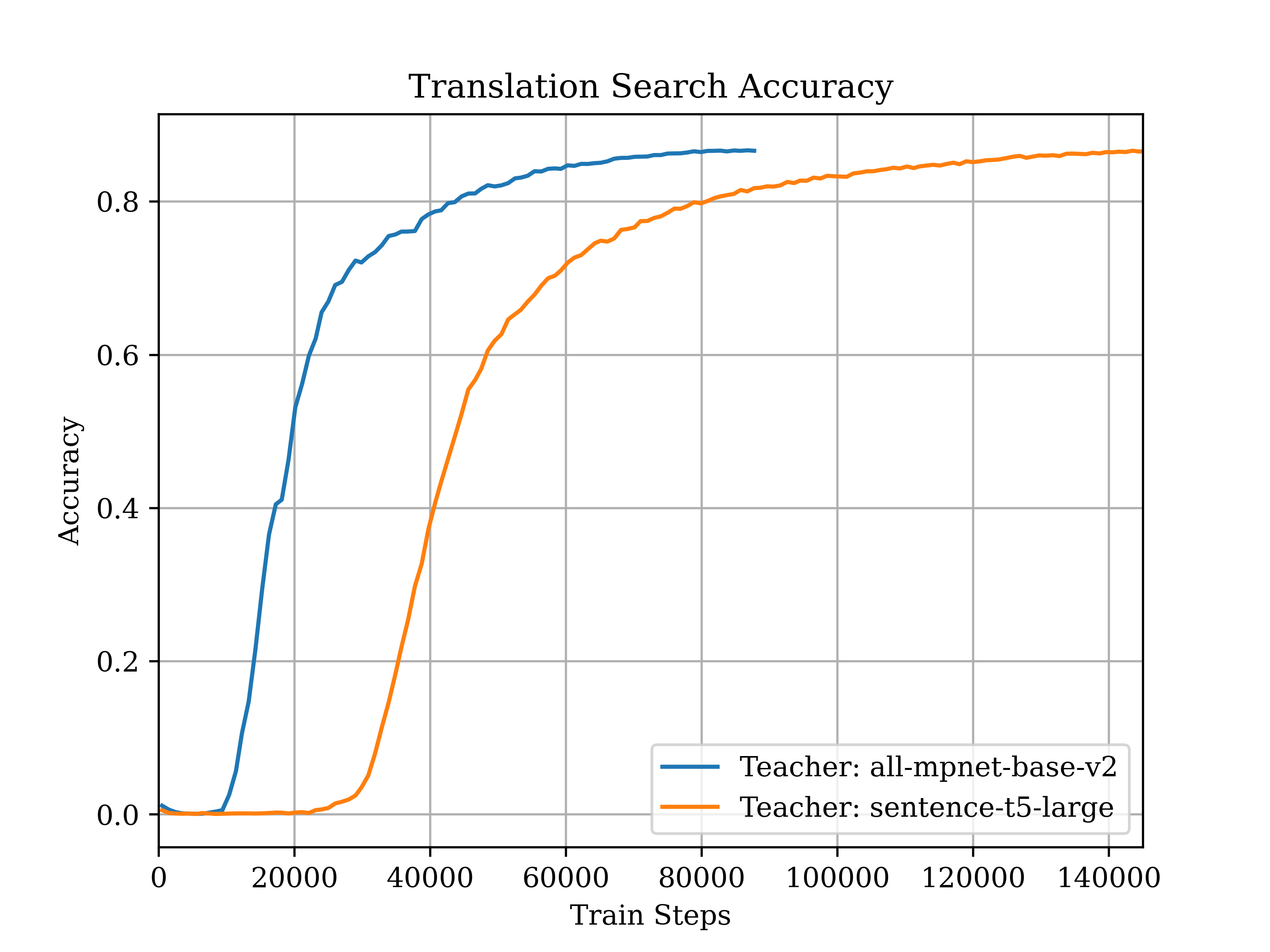}
	\caption{Translation search accuracy over training steps with \textsubscript{grc}XLM-R student model.}
	\label{fig:fig_acc}
\end{figure}

We compare \textsubscript{GRC}mBERT and \textsubscript{GRC}XLM-R as the student model $\hat{M}$.
We add a mean pooling layer and pair both student models with both teacher models (4 configurations)
and train all the student parameters. With \texttt{mpnet} as the teacher, we train for 15 epochs,
but with \texttt{st5} the student model took twice as long to converge (See Figure \ref{fig:fig_acc}),
so we train for 30 epochs. We use a batch size of 128, a max sequence length of 128 tokens, 2000 warmup steps, and a learning rate of 2e-5.
Every 500 training steps we measure STS performance as well as MSE loss and translation search
accuracy on 5k hold-out pairs, keeping the model with best average performance across these tasks.
Regardless of teacher model, \textsubscript{GRC}XLM-R took many more training steps
to converge than \textsubscript{GRC}mBERT and was prone to catestrophic forgetting, which
was alleviated by increasing the number of warmup steps.

We also experiment with training on parallel Modern Greek data from Wikipedia
for 3 epochs and then on Ancient Greek data for 15 epochs if \texttt{mpnet} is
the teacher and 6 and 30 epochs if \texttt{st5} is the teacher. Although
Modern Greek differs in many significant ways from Ancient Greek, training on this
data gives the model additional exposure to aspects of Greek
that have remained unchanged since antiquity, such as historical proper nouns.
All evaluations are reported with and without training on this additional data.

\subsection{Contrastive Learning}
\label{sec:simcse}

As a baseline against which to compare the models trained via the distillation method,
we also train sentence embedding models using \textit{Simple Contrastive Learning of Sentence Embeddings}
(SimCSE), the contrastive learning method introduced by \citet{gao-etal-2021-simcse}.
Contrastive learning pulls semantically-close neighbors together and pushes apart
non-neighbors, and has been shown to be effective for training multilingual sentence
embeddings \citep{gao-etal-2021-simcse,tan-etal-2023-multilingual}. In addition to using dropout as noise,
we use each Greek sentence and its English translation as positive pairs and
other pairs in the same batch as negatives.

We use the CLS token representation and train for a maximum of 10 epochs with a batch size of 82,
a max sequence length of 128 tokens, 2000 warmup steps, and a learning rate of 2e-5.
Every 500 training steps we measure performance on the STS evaluation and
translation search accuracy on the 5k hold-out pairs,
keeping the highest performing model. As above,
we also experiment with training on Modern Greek data for 3 epochs,
and then Ancient Greek data for 10 epochs.

\section{Training Data}

\subsection{Pre-training}

Our pre-training dataset consists of the Ancient Greek text from
the Perseus Digital Library\footnote{\url{https://github.com/PerseusDL/canonical-greekLit}}
and First1KGreek,\footnote{\url{https://github.com/OpenGreekAndLatin/First1KGreek}} which
are part of the Open Greek and Latin project.\footnote{\url{https://opengreekandlatin.org}}
Different documents containing the same Greek work were removed.
These sources contain approximately 32 million words of Ancient Greek text.
Although \citet{riemenschneiderExploringLargeLanguage2023} produced a much larger corpus
of Greek text (100+ million words) using additional sources, at the time of
writing their data is not publicly available.
Our smaller dataset is sufficient for our purposes,
as \citet{reimers-gurevych-2020-making} show
that even languages with little pre-training in a multilingual student
model can be effective targets for knowledge distillation.

This dataset consist of Greek texts spanning a thousand years, covering different dialects and time periods
of the language. We do not filter out any texts based on their dialect or time period.

In addition to the Greek text, we also collect all the English translations in the Open Greek and Latin
project to finetune our models with an additional 10 million words of historical English text.

\subsection{Preprocessing}

Following \citet{yamshchikov-etal-2022-bert} and \citet{singh-etal-2021-pilot},
we lowercase all the Greek text and strip diacritics, but keep punctuation. Although
diacritics contain important information for disambiguating between words that only differ
by breathing marks or accent marks, the correct word can usually be inferred from context.
The contextual nature of BERT models allows them to learn to use context to disambiguate.

\subsection{Parallel Data}

\paragraph{Human Aligned}
A portion of our parallel sentence dataset is taken from human aligned sources:
\begin{enumerate}
  \item Verses of the Greek New Testament with English translations (15k pairs),
  \item Verses of the Greek Septuagint with English translations (29k pairs),
  \item Verses of the Greek works of Flavius Josephus with English translations (15k pairs),
  \item Other minor sources: OPUS \citep{tiedemann-nygaard-2004-opus},
  Greek Learner Texts\footnote{\url{https://greek-learner-texts.org}},
  manually aligned passages from Perseus and First1KGreek (total 23k pairs).
\end{enumerate}

\paragraph{Translation Alignment}
The bulk of the parallel data is produced using translation alignment.
We take all the texts from our pre-training corpus that have English translations
and split them into sentences or sub-sentence segments (see Appendix \ref{sec:appendix_b}).
We then use a two-step process to align Greek sentences with their English translations.
First, we use Hunalign \citep{halacsyParallelCorporaMedium2007}, a sentence-length heuristic and dictionary-based alignment technique
on the translated texts.
This produced an initial dataset of approximately 150k parallel sentences
(including the human-aligned sources listed above).

Using this initial dataset, we trained a sentence embedding model with an aligned
vector space for English and Ancient Greek using SimCSE (See Section \ref{sec:simcse}).
Next, we use this model to align all the texts again, using a better alignment method introduced by \citet{liuBertalignImprovedWord2023},
dubbed Bertalign, which uses multilingual sentence embeddings to achieve state-of-the-art alignment quality.
If the Greek and English documents are already aligned by sections, we align the sentences in each section
individually. This increases alignment accuracy and makes it possible to keep the parts
of the document that have good alignments and to discard the rest. Otherwise, if no section
alignments exist, we run the aligner on the entire text.

We do not filter out multiple translations of the same Greek texts, since different translations
can have different nuances and word choices, with the hope that the resulting sentence
embeddings will be more robust to translation differences.

Finally, we remove all duplicate sentence pairs from the dataset and all pairs with very
short sentences ($<$5 characters). We also ensure that no sentence pairs from the
STS dataset (See Section \ref{sec:sts}) are included in the training data.
This results in approximately 380k sentence pairs after holding out 5k
pairs for evaluation purposes.

\paragraph{Modern Greek}
The Modern Greek (EL) sentence pairs from Wikipedia are taken from the OPUS project \citep{tiedemann-nygaard-2004-opus}.
We remove all duplicate pairs and pairs with very short sentences ($<$10 characters), resulting
in approximately 800k sentence pairs.
This dataset contains a rich and diverse set of topics, including historical topics
which will hopefully transfer to the Ancient Greek models.
We compare all the models with and without training on this data.

\section{Evaluations}

\subsection{Translation Similarity Search}

The first measure of the quality of the sentence embeddings is each model's
accuracy at choosing the correct English translation for each Ancient Greek sentence
from the 5k hold-out pairs.
The score is computed as the percentage of sentence pairs for which the
embedding of source sentence \(s_{i}\) has the closest cosine similarity to
the embedding of translated sentence \(t_{i}\) out of all the target sentences.
The accuracy is computed in both directions and averaged.
The results are reported in Table \ref{table:eval_translation}.

\begin{table}[ht]
	\small
	\centering
	\begin{tabular}{lr}
		\textbf{Model} & \textbf{Accuracy} \\
		\toprule
		\textit{SimCSE} \\
		\toprule
		\textsubscript{GRC}mBERT (GRC)                 & 95.92 \\
		\textsubscript{GRC}mBERT (EL,GRC)              & 96.09 \\
		\textsubscript{GRC}XLM-R (GRC)                 & 95.86 \\
		\textsubscript{GRC}XLM-R (EL,GRC)              & \textbf{96.64} \\
		\midrule
		\multicolumn{2}{l}{\textit{Teacher:} \texttt{sentence-t5-large}} \\
		\toprule
		\textsubscript{GRC}mBERT (GRC)                 & 87.78 \\
		\textsubscript{GRC}mBERT (EL,GRC)              & 90.80 \\
		\textsubscript{GRC}XLM-R (GRC)                 & 87.02 \\
		\textsubscript{GRC}XLM-R (EL,GRC)              & 91.60 \\
		\midrule
		\multicolumn{2}{l}{\textit{Teacher:} \texttt{all-mpnet-base-v2}} \\
		\toprule
		\textsubscript{GRC}mBERT (GRC)                 & 87.77 \\
		\textsubscript{GRC}mBERT (EL,GRC)              & 89.15 \\
		\textsubscript{GRC}XLM-R (GRC)                 & 86.48 \\
		\textsubscript{GRC}XLM-R (EL,GRC)              & 90.12 \\
		\bottomrule
	\end{tabular}
	\caption{Translation similarity search accuracy. Best result is bolded.}
	\label{table:eval_translation}
\end{table}

The SimCSE models perform on this task better than the distillation models, which
is not surprising since they specifically trained to maximize the cosine similarity
between translation pairs and minimize similarity between non-pairs. There is no significant difference in the performance
between the two base models.
All the models performed better when first trained on Modern Greek before Ancient Greek.

\subsection{Semantic Textual Similarity}
\label{sec:sts}

\begin{table}[htb]
	\small
	\centering
	\begin{tabular}{p{6.4cm}r}
		\multicolumn{2}{l}{
			\begin{tabular}{@{}p{6.12cm}r@{}}
				\textbf{Sentence Pair} & \textbf{Score} \\
			\end{tabular}
		} \\
		\midrule
		\textgreek{Στωικοὶ ἀποφαίνονται σφαιροειδῆ τὸν κόσμον.} \\
		Stoics declare the world to be spherical. & 0.9 \\
		\textgreek{Στωικός νομίζει ὅτι ἡ γῆ σφαίρα ἐστιν.} \\
		A Stoic thinks that the earth is a sphere.  \\
		\midrule
		\textgreek{ἐπὶ δὲ τοῦ ὄρους τῇ ἄκρᾳ Διός ἐστιν ναός.} \\
		On the top of the mountain is a temple of Zeus. & 0.8 \\
		\textgreek{ὁ Ζεὺς οἰκεῖ ἐπὶ τὰ ὄρη ἐν Ὀλύμπῳ.} \\
		Zeus dwells on the mountains in Olympus.  \\
		\midrule
		\textgreek{Τὰ παιδία παίζουσιν ἐν τῇ ἀμμουδιᾷ.} \\
		The children are playing in the sand. & 0.5 \\
		\textgreek{Τὰ παιδία ἀναπαύονται ἐν τῷ κήπῳ.} \\
		The children rest in the garden. \\
		\midrule
		\textgreek{Σωκράτης εἶδεν ἓξ βόας.} \\
		Socrates saw six cows. & 0.1 \\
		\textgreek{Ῥώμουλος εἶδεν ἓξ οἰωνοὺς ὄρνιθας.} \\
		Romulus saw six birds of omen. \\
		\bottomrule
	\end{tabular}
	\caption{Example pairs from STS evaluation dataset. Scores are examples and not actual scores.}
	\label{table:sts_examples}
\end{table}

\begin{table*}[t]
	\small
	\centering
	\begin{tabular}{lcccc}
		\textbf{Model}     & \textbf{GRC$\leftrightarrow$GRC} & \textbf{EN$\leftrightarrow$EN} & \textbf{GRC$\leftrightarrow$EN} & \textbf{Average} \\
		\toprule
		\textit{SimCSE} \\
		\toprule
		\textsubscript{GRC}mBERT (GRC)   & 75.68 & 77.58 & 76.30 & 76.52 \\
		\textsubscript{GRC}mBERT (EL,GRC)   & 74.85 & 78.30 & 76.40 & 76.52 \\
		\textsubscript{GRC}XLM-R (GRC)   & 77.83 & 78.82 & 77.21 & 77.95 \\
		\textsubscript{GRC}XLM-R (EL,GRC)   & 78.27 & 79.11 & 77.76 & 78.38 \\
		\midrule
		\multicolumn{4}{l}{\textit{Teacher:} \texttt{sentence-t5-large}} \\
		\toprule
		\textsubscript{GRC}mBERT (GRC)  & 82.17 & 87.54 & 84.02 & 84.58 \\
		\textsubscript{GRC}mBERT (EL,GRC)  & 84.84 & \textbf{89.33} & \textbf{86.37} & \textbf{86.84} \\
		\textsubscript{GRC}XLM-R (GRC)  & 82.37 & 85.37 & 82.56 & 83.43 \\
		\textsubscript{GRC}XLM-R (EL,GRC)  & 84.88 & 88.37 & 85.45 & 86.24 \\
		\midrule
		\multicolumn{4}{l}{\textit{Teacher:} \texttt{all-mpnet-base-v2}} \\
		\toprule
		\textsubscript{GRC}mBERT (GRC)  & 82.30 & 87.60 & 84.68 & 84.86 \\
		\textsubscript{GRC}mBERT (EL,GRC)  & 84.84 & 88.77 & 86.28 & 86.63 \\
		\textsubscript{GRC}XLM-R (GRC)  & 83.80 & 87.07 & 84.53 & 85.13 \\
		\textsubscript{GRC}XLM-R (EL,GRC)  & \textbf{85.18} & 88.24 & 85.92 & 86.45 \\
		\bottomrule
	\end{tabular}
	\caption{Spearman rank correlation \(\rho\)
		between the cosine similarity of sentence embeddings and gold
		labels for STS dataset. Scores are reported as \(\rho \times 100\).
		Best results are bolded.
		There are twice as many \textit{GRC-EN} pairs as \textit{GRC-GRC} pairs so their scores are not directly comparable.
		}
	\label{table:eval_sts}
\end{table*}

We compiled a dataset of Ancient Greek sentence pairs with gold scores to
measure semantic textual similarity in the range [0,1],
with 0 representing completely unrelated meaning, and 1 representing full semantic equivalence.
Each sentence was given a corresponding English translation to allow for cross-lingual evaluation
(See Table \ref{table:sts_examples}).

\begin{figure}[h]
	\centering
	\includegraphics[width=7cm]{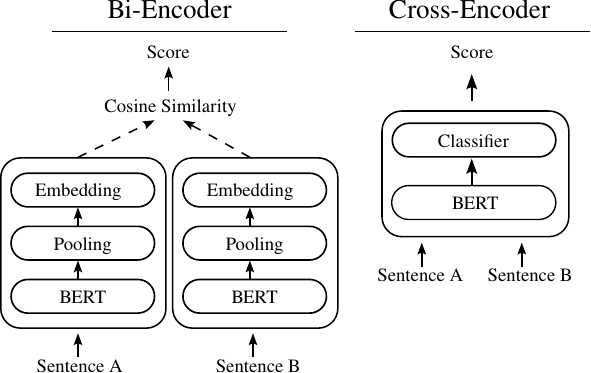}
	\caption{We use a Cross-Encoder (right) to produce STS gold scores which are used to evaluate our
	sentence embedding models, which are Bi-Encoders (left).}
	\label{fig:bi_vs_cross}
\end{figure}

The gold scores for STS datasets are typically produced by averaging the scores from many human
annotators. However, for Ancient Greek it is difficult to find enough
annotators to produce high quality gold scores. Our solution is to use a
Cross-Encoder\footnote{\url{https://huggingface.co/cross-encoder/stsb-roberta-base}}
to produce the gold scores based on the English translations of each pair.
A Cross-Encoder takes two sentences as input and produces a similarity score in the
range $[0,1]$ without the need to encode
the semantic properties of each sentence into a vector, and therefore performs better
than cosine similarity between embeddings
(See Figure \ref{fig:bi_vs_cross}).
With this setup, we measure how closely each model can match the performance
of the English Cross-Encoder.
The accuracy of this method depends on how closely the English translations match the meaning
of the Greek sentences. Therefore the English translations are reviewed by an expert to
ensure that they are literal and accurate translations of the Greek text.

Due to the need to manually verify the translations for each pair, the STS dataset is relatively small.
The dataset consists of 165 Ancient Greek sentences pairs, each having an English translation:
\(((a_{GRC},a_{EN}), (b_{GRC},b_{EN}))\).
The GRC$\leftrightarrow$EN comparison can be performed two ways:
\(a_{GRC} \leftrightarrow b_{EN}\) and \(a_{EN} \leftrightarrow b_{GRC}\) for a total of 330
GRC$\leftrightarrow$EN comparisons, 165 GRC$\leftrightarrow$GRC comparisons,
and 165 EN$\leftrightarrow$EN comparisons.

The score for each model is computed as Spearman correlation between gold scores and the cosine similarities
between the sentence embeddings. The results are reported in Table \ref{table:eval_sts}.

The models trained via knowledge distillation significantly outperform the SimCSE models,
showing that they have inherited the properties of the teacher models for
STS tasks. The models with the \texttt{st5} teacher have a small lead,
which is expected since \texttt{st5} was trained for STS tasks.
All the models improve slightly when first trained on Modern Greek before Ancient Greek.

\subsection{Semantic Retrieval}

Sentence embeddings can be used for semantic retrieval tasks
by ranking a set of passage embeddings by cosine similarity
with a query embedding. Performing this process with our models on the Greek sentences
in the Perseus and First1KGreek corpora yields promising results. For example,
the following query is correctly answered by several passages in the top 10
highest ranked passages:

\textbf{Query:} ``Was Aristotle a student of Plato?''
\begin{itemize}
	\small
	\item \textgreek{Ἀριστοτέλης Πλάτωνος μαθητής· οὗτος τὴν διαλεκτικὴν συνεστήσατο.} - Hyppolytus of Rome \newline
	Aristotle, a disciple of Plato — He established dialectics.
	\item \textgreek{ἀλλὰ καὶ τοῖς Πλάτωνος ἐγκαλέσαι ἄν τις δόγμασι δι' Ἀριστοτέλην, ἀποφοιτήσαντα τῆς διατριβῆς αὐτοῦ ἐν καινοτομίαις.} - Origen \newline
	But someone could also challenge certain doctrines of Plato through Aristotle, who, upon completing his studies, departed from his teachings with innovations.
	\item \textgreek{παρὰ Πλάτωνι Ἀριστοτέλης φιλοσοφήσας μετελθὼν εἰς τὸ Λύκειον κτίζει τὴν Περιπατητικὴν αἵρεσιν.} - Clement of Alexandria \newline
	After studying philosophy under Plato, Aristotle, having come to the Lyceum, founded the Peripatetic school.
\end{itemize}

To quantify the performance of each model for semantic retrieval, we compile
a dataset of 40k Greek passages from the Perseus and First1KGreek corpora.
We then produce 100 English queries (in the form of both phrases and questions) and associate
them with relevant passages.
We measure recall and mean average precision (mAP) for each model.
The scores are reported in Table \ref{table:eval_sr}.

The SimCSE models perform poorly, which is expected since they were not trained
for retrieval tasks. The models with the \texttt{mpnet} teacher,
which was trained for semantic search, score highest by a large margin.
The models with the \texttt{st5} teacher, which was trained for semantic textual similarity
tasks, perform better than the SimCSE models but worse than the \texttt{mpnet} models.
The models generally perform much better when trained on Modern Greek.
Perhaps this is because many of the queries involve
proper nouns for which Modern Greek data gave additional training examples, or
perhaps the student models benefited from the additional English examples to learn the vector space
properties of the teacher.
The \textsubscript{GRC}mBERT models consistently perform better than \textsubscript{GRC}XLM-R models.

Overall performance on this task was rather poor even for the best models.
An analysis of the top ranked passages for each query revealed that passages
about related topics often ranked above the desired passages. In particular, it often
confused proper names, e.g. preferring passages about other philosophers
for queries about Plato.

\begin{table}[htb]
	\centering
	\small
	\begin{tabular}{lrrr}
		\textbf{Model}     & \textbf{Recall@10} & \textbf{mAP@20} \\
		\toprule
		\textit{SimCSE} \\
		\toprule
		\textsubscript{GRC}mBERT (GRC)   &  26.61  & 15.33 \\
		\textsubscript{GRC}mBERT (EL,GRC)   &  18.08  & 10.84 \\
		\textsubscript{GRC}XLM-R (GRC)   &  21.50  & 9.86 \\
		\textsubscript{GRC}XLM-R (EL,GRC)   &  29.56  & 15.08 \\
		\midrule
		\multicolumn{3}{l}{\textit{Teacher:} \texttt{sentence-t5-large}} \\
		\toprule
		\textsubscript{GRC}mBERT (GRC)  &  41.34  & 25.37 \\
		\textsubscript{GRC}mBERT (EL,GRC)  &  49.63  & 36.17 \\
		\textsubscript{GRC}XLM-R (GRC)  &  34.88  & 20.07 \\
		\textsubscript{GRC}XLM-R (EL,GRC)  &  47.07  & 31.31 \\
		\midrule
		\multicolumn{3}{l}{\textit{Teacher:} \texttt{all-mpnet-base-v2}} \\
		\toprule
		\textsubscript{GRC}mBERT (GRC)  &  63.60  & 44.97 \\
		\textsubscript{GRC}mBERT (EL,GRC)  &  \textbf{69.87}  & \textbf{53.00} \\
		\textsubscript{GRC}XLM-R (GRC)  &  53.84  & 36.42 \\
		\textsubscript{GRC}XLM-R (EL,GRC)  &  60.13  & 44.36 \\
		\bottomrule
	\end{tabular}
	\caption{Recall@10 and mAP@20 scores for English search queries and Ancient Greek passages. Best results are bolded.
		}
	\label{table:eval_sr}
\end{table}

\subsection{Translation Bias}
\label{sec:bias}

\begin{table*}[htb]
	\centering
	\small
	\begin{threeparttable}
	\begin{tabular}{lrrrrrrrrr|r}
		\textbf{Model} & \textbf{KJV} & \textbf{NKJV}\tnote{*} & \textbf{NASB} & \textbf{ESV} & \textbf{RSV} & \textbf{NET}\tnote{*} & \textbf{NIV} & \textbf{NLT} & \textbf{MSG} & \textbf{Avg. Emb.} \\
		\midrule
		\textit{SimCSE} \\
		\toprule
		\textsubscript{GRC}mBERT (GRC)    &  32.59 & 36.56 & \textbf{39.97} & 32.17 & 29.28 & 25.91 & 20.32 & 12.92 & 11.14 & \underline{52.04} \\
		\textsubscript{GRC}mBERT (EL,GRC) &  33.27 & 36.05 & \textbf{40.79} & 31.97 & 28.74 & 26.17 & 20.15 & 12.79 & 11.21 & \underline{51.75} \\
		\textsubscript{GRC}XLM-R (GRC)    &  35.78 & 37.85 & \textbf{38.17} & 32.15 & 29.76 & 26.03 & 20.58 & 13.14 & 11.32 & \underline{48.11} \\
		\textsubscript{GRC}XLM-R (EL,GRC) &  35.34 & 36.74 & \textbf{38.02} & 32.96 & 30.06 & 25.76 & 20.75 & 13.09 & 11.25 & \underline{48.93} \\
		\midrule
		\multicolumn{3}{l}{\textit{Teacher:} \texttt{sentence-t5-large}} \\
		\toprule
		\textsubscript{GRC}mBERT (GRC)    &  29.63 & \textbf{30.70} & 30.13 & 27.81 & 25.90 & 23.05 & 20.09 & 14.43 & 12.49 & \underline{78.66} \\
		\textsubscript{GRC}mBERT (EL,GRC) &  28.73 & \textbf{30.66} & 29.98 & 28.02 & 25.82 & 23.39 & 19.70 & 13.93 & 12.24 & \underline{80.42} \\
		\textsubscript{GRC}XLM-R (GRC)    &  \textbf{31.82} & 29.70 & 29.14 & 28.10 & 26.39 & 23.82 & 20.38 & 14.24 & 12.90 & \underline{76.40}\\
		\textsubscript{GRC}XLM-R (EL,GRC) &  \textbf{30.41} & 30.01 & 29.08 & 28.35 & 26.82 & 23.61 & 19.75 & 13.68 & 12.36 & \underline{78.82} \\
		\midrule
		\multicolumn{3}{l}{\textit{Teacher:} \texttt{all-mpnet-base-v2}} \\
		\toprule
		\textsubscript{GRC}mBERT (GRC)    &  \textbf{31.76} & 31.15 & 29.93 & 30.04 & 28.94 & 23.33 & 19.52 & 13.93 & 11.98 & \underline{72.32} \\
		\textsubscript{GRC}mBERT (EL,GRC) &  30.42 & \textbf{31.20} & 30.37 & 30.35 & 28.68 & 23.51 & 19.53 & 13.76 & 11.81 & \underline{73.26} \\
		\textsubscript{GRC}XLM-R (GRC)    &  \textbf{37.25} & 31.31 & 29.91 & 30.00 & 29.42 & 23.32 & 19.67 & 13.97 & 12.32 & \underline{65.74} \\
		\textsubscript{GRC}XLM-R (EL,GRC) &  \textbf{33.08} & 31.21 & 29.64 & 30.25 & 29.59 & 23.55 & 19.32 & 13.61 & 11.89 & \underline{70.76} \\
		\bottomrule
	\end{tabular}
	\begin{tablenotes}
		\item [*] Verses from the NET and NKJV were included in parallel training data.
	\end{tablenotes}
	\end{threeparttable}
	\caption{Mean Reciprocal Rank (MRR) $\times 100$ of cosine similarity between Greek verses of the New Testament and English
		translations, as well as MRR of per-verse averaged embedding of all the translations.
		Highest translation MRR for each model is bolded.
		MRR of averaged embedding is underlined if it is higher than any of the translations.
		}
	\label{table:eval_all_nt}
\end{table*}

To determine whether the models are biased towards certain translation styles,
especially those included in the training set, a text with many different
translations is needed. The New Testament is a good candidate for this,
since many translations exist in different styles
and eras of the English language. We take nine New Testament translations,
ranging from literal (NASB), archaic (KJV), and paraphrase (MSG),
all fully aligned at the verse level (7654 verses).
There are no other Greek texts that we are aware of that have
this many translations available for comparison.
We generate embeddings for each verse from the Greek text and the
translations. We also generate an ``average'' translation for each verse
by averaging the embeddings of all the English translations.
We take the cosine similarity between the Greek embedding and
each translation and use it to compute the Mean Reciprocal Rank (MRR)
across all verses, for each model:
\[
	MRR=\frac{1}{|T|}\sum_{v \in T} \frac{1}{rank_{v}}
\]
where $T$ is a set of verses in a translation
and $rank_{v}$ is the rank of the translation for verse $v$.
The results are reported in Table \ref{table:eval_all_nt}.

The literal translations score highest, and the score decreases the more
non-literal the translations become, with the MSG translation having the lowest score. Surprisingly, the archaic KJV translation
ranks highly, which is likely due to a high quantity of archaic English text
in the training data. This suggests that the models are slightly biased to this older
English translation style.
Verses from two of the translations (NKJV and NET) were included in the training data.
Despite being in the training data, there does not appear to be bias to the NET
since it consistently ranks lower than other translations. The NKJV ranks highly,
but does not consistently outrank other literal translations.
Interestingly, the average embedding of all the translations ranked
highest by a significant margin.

\section{Discussion and Future Work}

Overall, the base models mBERT and XLM-R
performed similarly except for the semantic retrieval task where
the mBERT-derived models have a sizeable lead.
The reason for this is unclear,
since these models have different tokenizers, parameter counts, and vocabularies.
It is also unclear how much the pre-training process affects the results.
An area of future research would be to investigate the effect of student model
architecture, tokenizer, and pre-training on the ability of the student model to learn from
the teacher model.

The main limitation of using multilingual knowledge distillation to train sentence embedding models
is that the embeddings produced are almost entirely derived from English translations,
which could be undesirable if the goal is to study Ancient Greek text without
any prior translator's interpretation. Furthermore, the student model can never fully replicate the
performance of the teacher model when transfering to another language, since translated sentences
are often not entirely semantically equivalent to their source sentences, especially
when removed from the original context via translation alignment.

Although contamination from
modern languages is not a big concern for the tasks in this paper, there
could be issues of anachronisms when searching Ancient Greek texts with
English. Furthermore, using texts from such a long chronological period
of the Greek language could introduce additional lexical polysemy as Greek
words changed in meaning over time. This could explain why the averaged
embedding of many translations had a higher MRR than any individual
translation source in Table \ref{table:eval_all_nt}, since the combination
of many translations represents a higher degree of polysemy.
In future work, such historical polysemy could be measured by sampling
translations of words from texts of different historical periods.
This could help to determine whether the high MRR of the averaged embedding
is a useful result or simply an artifact of a potentially high amount of polysemy
in the training data.

\section{Conclusion}

In this paper, we have shown that multilingual knowledge distillation
is an effective approach for training sentence embedding models
for Ancient Greek, in spite of the lack of available
training data compared to modern, high-resource languages. 
In addition, we have produced a new dataset of parallel Ancient Greek
and English sentences as well as evaluation datasets for translation
search, semantic textual similarity, and semantic retrieval, which
we make publicly available.

\bibliography{anthology,custom}

\appendix

\section{Appendix: Training Details}
\label{sec:appendix_a}

\begin{table}[htbp]
	\centering
	\small
	\begin{tabular}{lll}
		\textbf{Parameter} & \textbf{\textsubscript{GRC}mBERT} & \textbf{\textsubscript{GRC}XLM-R} \\
		\midrule
    Batch Size & 140 & 128 \\
    Learning Rate & 2e-5 & 2e-5 \\
    LR Scheduler & linear & linear \\
    Epochs & 10 & 10 \\
    Warmup Steps & 2000 & 2000 \\
    Mask Percentage & 15\% & 15\% \\
    \bottomrule
	\end{tabular}
	\caption{Pre-training hyperparameters}
\end{table}

\begin{table}[htbp]
	\centering
	\small
	\begin{tabular}{lll}
		\textbf{Parameter} & \textbf{\textsubscript{GRC}mBERT} & \textbf{\textsubscript{GRC}XLM-R} \\
		\midrule
		Batch Size & 128 & 128 \\
		Learning Rate & 2e-5 & 2e-5 \\
		LR Scheduler & linear & linear \\
		Max Seq. Length & 128 & 128 \\
		Pooling & mean & mean \\
		Embedding Dim. & 768 & 768 \\
		\bottomrule
		\multicolumn{3}{l}{\textit{Teacher:} \texttt{all-mpnet-base-v2}} \\
		\toprule
		Epochs (GRC) & 15 & 15 \\
		Epochs (EL) & 3 & 3 \\
		GRC Warmup Steps & 2000 & 2000 \\
		EL Warmup Steps & 2000 & 8000 \\
		\bottomrule
		\multicolumn{3}{l}{\textit{Teacher:} \texttt{sentence-t5-large}} \\
		\toprule
		Epochs (GRC) & 30 & 30 \\
		Epochs (EL) & 6 & 6 \\
		GRC Warmup Steps & 2000 & 2000 \\
		EL Warmup Steps & 2000 & 2000 \\
		\bottomrule
	\end{tabular}
	\caption{Knowledge distillation hyperparameters}
\end{table}

\begin{table}[htbp]
	\centering
	\small
	\begin{tabular}{lll}
		\textbf{Parameter} & \textbf{\textsubscript{GRC}mBERT} & \textbf{\textsubscript{GRC}XLM-R} \\
		\midrule
		Batch Size & 82 & 82 \\
		Learning Rate & 2e-5 & 2e-5 \\
		LR Scheduler & linear & linear \\
		Warmup Steps & 2000 & 2000 \\
		Max Seq. Length & 128 & 128 \\
		Epochs (GRC) & 10 & 10 \\
		Epochs (EL) & 3 & 3 \\
		Pooling & CLS & CLS \\
		Embedding Dim. & 768 & 768 \\
		\bottomrule
	\end{tabular}
	\caption{SimCSE hyperparameters}
\end{table}

\section{Appendix: Sentence Segmentation}
\label{sec:appendix_b}

For translation alignment, it is not necessary that each segment be a sentence,
since the alignment process can handle 1-many, many-1 or many-to-many relations.
The Greek texts in our corpus contain punctuation, so we segment them by 
period (.), question mark (;), and raised dot (·). Some of the Greek texts
use a colon instead of a raised dot, and in these cases we treat colons as raised
dots. For the English texts we first segment using the NLTK
sentence tokenizer \citep{birdNaturalLanguageProcessing2009} then further subdivide
these segments by semicolon (;) and colon (:).

\end{document}